\documentclass{article}
\usepackage{iclr2027_conference,times}
\usepackage[utf8]{inputenc}
\usepackage[T1]{fontenc}
\usepackage{amsmath,amssymb,mathtools}
\usepackage{booktabs,array,tabularx}
\usepackage{float}
\usepackage{xcolor}
\usepackage{microtype}
\usepackage{url}
\usepackage{hyperref}
\usepackage{cleveref}
\definecolor{accent}{RGB}{36,82,150}
\hypersetup{colorlinks=true,hypertexnames=false,linkcolor=accent,citecolor=accent,urlcolor=accent,
 pdftitle={From Transfer Probes to Expert Grouping: Evidence from LoRA Post-Training},
 pdfsubject={Technical report formatted with the ICLR 2027 conference style}}
\newcolumntype{Y}{>{\raggedright\arraybackslash}X}

\title{From Transfer Probes to Expert Grouping:\\Evidence from LoRA Post-Training}
\author{Xueyang Wu}
\iclrfinalcopy
\begin{document}
\maketitle
\lhead{Technical report -- ICLR 2027 layout (not a conference paper)}
\thispagestyle{fancy}
\begin{abstract}
A platform that post-trains one base model for many business lines often has more lines than experts it can maintain, so lines must share experts, yet lines seldom come with a taxonomy saying which should share. We ask whether transfer probes (train on one source, measure paired changes on all sources) can discover a grouping when the standard one is unknown. To test probes against a reference, we use seven public sources that do have a label tree (five mathematics subfields and two kinds of programming, standing in for business lines) and treat the tree's grouping as a label reference that the evaluator holds and the method does not. With four LoRA experts under fixed per-source exposure and per-expert budgets, we compare a purely probe-derived grouping and an uninformed uniform default (whose tied routing serves every line from one expert) against this reference. Trained once for one interval from a shared warm-up state and routed by a rule fixed by each allocation, neither reached the prespecified practical gain of 2 percentage points, and both had lower overall utility than the reference ($-2.43$ and $-2.54$ pp; 95\% intervals below zero). In a post hoc comparison, the probe-derived grouping was indistinguishable from the uniform default ($+0.11$ pp, $[-1.25,+1.54]$). In fixed-budget LoRA post-training, a grouping built from initial-state single-source probes thus did not deliver the advantage predicted from the probes once carried to a later training state, a finite budget, and specified routing.
\end{abstract}

\noindent{\footnotesize\textit{Technical-report status.} This document uses ICLR 2027 formatting but is not an ICLR submission or a peer-reviewed conference paper. Human authors made the research decisions and interpret the experimental results; AI assistance is disclosed below. Code and evidence repository: \href{https://github.com/CodePothunter/Reproduction-for-From-Transfer-Probes-to-Expert-Grouping}{GitHub (CodePothunter)}. The linked repository is separate from this manuscript source archive.}\par\medskip

\section{Introduction}\label{sec:intro}

Consider a platform that post-trains one base model for many business lines, such as different products, regions, or internal functions. Every request carries its line label, so serving can route by line. Low-rank adapters make each expert cheap to train \citep{hu2022lora}, and multi-adapter serving systems make many of them cheap to host \citep{sheng2024slora,chen2024punica}; yet each deployed expert must still be trained, validated, and maintained, so a platform may budget fewer experts than it has lines. Some lines must then share an expert, and the platform has to decide which.

If the lines came with a trusted similarity tree, a natural choice would be to cut the tree into as many groups as there are experts. Such a tree rarely exists: business lines are defined by organization and product, not by which lines help each other in training, so the standard grouping is unknown. A tempting substitute is to measure it. A \emph{transfer probe} trains an adapter on one source and records the paired change in accuracy on every source; a table of probes says, at the probed state, whether learning A carries B along. The question of this report is whether such probes can discover a grouping: whether a grouping built from probes matches a good reference grouping, or at least beats an allocation that uses no information.

Answering this requires a grouping to compare against, so we test probes where a label tree \emph{does} exist before anyone relies on them where it does not. Our seven sources, Integral and Differential Calculus, Abstract and Linear Algebra, and Statistics from DeepMath-103K \citep{he2025deepmath}, and algorithmic and generic programming problems from OpenCodeInstruct \citep{ahmad2025opencodeinstruct}, stand in for business lines. Their tree (mathematics splits into calculus, algebra, and Statistics; code into its two kinds) yields a reference grouping that we call the \emph{label reference} C0: privileged information that the evaluator holds and a probe-based method does not. We do not assume C0 is the best grouping; it is the grouping a practitioner holding the tree would choose. With four experts the tree has exactly one cut into four groups, so C0 is fixed by the tree and budget balancing alone, and no development response data influenced it. The sources were chosen so that label structure and response structure could disagree, and a response model fitted to development probes predicted that a grouping different from the tree's would score higher (Section~\ref{sec:setting}). The testbed therefore leaves room for probes to win, although the predicted margin, under one percentage point, was smaller than the 2-point gain the comparison was designed to detect (Section~\ref{sec:probes}).

The probes suggested structure that the tree does not encode. Each probe trains a freshly initialized adapter on the base model (Qwen3.5-2B-Base; \citealp{qwen35}) with 16 blocks of a single source and scores paired correctness on an audit split of 80 problems per source. Training on each of the four other mathematics sources moved Statistics upward ($+111/-36$ discordant items pooled over the four donors and two training-stream shuffles; descriptive), whereas the tree gives Statistics an expert of its own. Individual edges were less stable: under a Bonferroni correction over the 49 cells, no cell passed in the first shuffle and one passed in the second, so no edge passed in both. Against C0 we compared two groupings that can be built without the tree. The uninformed default C1 spreads every source evenly over the four experts; because all experts then tie, the routing rule serves every source from one of them. The probe-derived grouping C2 is the allocation that maximizes the utility predicted by an additive model of the first shuffle's probe tensor; it is obtained from transportation linear programs over the continuous relaxation of the feasible allocations, its optimal vertex is already integral in blocks, and it uses nothing from the tree. It serves Linear Algebra together with Statistics and Integral Calculus together with Abstract Algebra, departing from the tree within mathematics while keeping the two code sources together. C0, C1, and C2 were each trained once for one allocation interval from a shared warm-up state W, under identical per-source token exposure and per-expert budgets, and each was served by a routing rule computed from its own block allocation. All were scored on the same 400 test problems per source (for algorithmic programming, mostly problems from the external MBPP+ benchmark; Section~\ref{sec:setting}), two samples each, with jointly paired bootstrap intervals. Table~\ref{tab:state-evidence} records which evidence comes from which training state.

Under the W starting state, a single fixed-budget interval, and the specified block-matrix routing, neither candidate showed the prespecified practical gain of 2 percentage points (pp), and each had lower overall utility than C0 (Table~\ref{tab:results}): $-2.54$ $[-4.04,-1.12]$ pp for C1 and $-2.43$ $[-3.84,-1.14]$ pp for C2. A 2 pp gain is not supported and a shortfall is, since both upper bounds are below zero; whether the shortfall itself reaches 2 pp is not resolved. In a comparison added after observing these results, C2 was indistinguishable from C1: $+0.11$ pp, $[-1.25,+1.54]$, an interval that excludes differences of 2 pp in either direction. Because utility is linear in traffic weights, no reweighting of the lines gives either candidate a point-estimate advantage over C0 above 0.875 pp; per-line intervals nonetheless remain wide enough that a large gain for traffic concentrated on a single line cannot be excluded (Table~\ref{tab:weightfree}). Evaluated afterwards under the executed routing, the additive model from which C2 was derived had placed it $+2.01$ pp above C0, and the same model with every transfer term removed predicted no difference ($+0.02$ pp); within that model, the disappearance of probed transfer by state W does not by itself account for the shortfall. We cannot attribute the gap to a single mechanism: between probe and comparison the training state (fresh adapter versus W), the item bank (audit versus test, and for algorithmic programming also the benchmark), the dose (16 blocks of one source versus at most 8 per source in an interval), the training mix (one source versus several sources per expert), and the routing all changed, and C2 was selected by maximizing the very model that predicted its gain (Section~\ref{sec:lessons}).

That task relationships depend on training state, and can differ between transfer and joint training, is known from multi-task learning in vision \citep{standley2020which,fifty2021efficiently}; we do not claim the gap between transfer and grouping as a new phenomenon. Our contributions are:
\begin{itemize}
\item A testbed for checking probe-based grouping in a setting that has a label tree: a label reference that states its information source (here, only the tree and budget balancing), budgets that hold every source's exposure and every expert's share fixed, and a state--evidence map that separates where probes were measured, where candidates were constructed, where they were trained, and where they were scored (Section~\ref{sec:setting}, Table~\ref{tab:state-evidence}).
\item An executed, paired comparison in LoRA post-training: with fixed source exposure and expert budgets, a grouping constructed from initial-state single-source probes did not deliver the advantage predicted from the probes once carried to the W state, a finite budget, and specified routing; it scored below the label reference and was indistinguishable from the uninformed default (Tables~\ref{tab:results} and~\ref{tab:weightfree}).
\item Operational lessons backed by this run: check the probe state against the deployment state, record training allocation and serving routing separately, size evaluations for sampled decoding, under which two samples from the same model disagreed on 20--34\% of test items per source, and state what information a label reference uses (Section~\ref{sec:lessons}).
\item An evidence package with the code, per-item scores, and analysis outputs behind the reported numbers (Appendix~\ref{app:provenance}), and a list of open questions, each with the data it needs and a rough compute cost extrapolated from this run (Section~\ref{sec:open}).
\end{itemize}
The enterprise setting serves only as motivation: the experiments use public data, one starting state, one allocation interval, and one training run per candidate.

\section{Related work}\label{sec:related}

\paragraph{Which tasks should be trained together.}
\citet{standley2020which} ask which tasks should share a network under an inference budget, and select groupings by training candidate multi-task networks. Their findings cut both ways for probes. On the cautionary side, they find no correlation between multi-task affinity (whether two tasks help each other when trained together) and transfer-learning affinity in any of their four vision settings (Pearson $r$ between $-0.24$ and $-0.12$), and multi-task affinity itself changed with network capacity and training-set size. On the encouraging side, cheaper predictors worked in their settings: a network's validation loss after a pass through 20\% of the training data showed a moderately high correlation with its final loss (Pearson $r=0.49$ in one setting; their Section~5.3.1), and predicting a larger group's per-task losses by averaging those of fully trained two-task networks (their Section~5.3.2), which ignores complex task interactions, chose groupings that performed very similarly to those selected with every candidate fully trained, although with five tasks more than half of the candidates still had to be trained. \citet{fifty2021efficiently} (TAG) measure inter-task affinity within a single training run as the change in one task's loss after a lookahead gradient step on another, and choose groups that maximize it. Their analysis (Section~4.3) gives a simple quadratic counterexample in which higher affinity does not imply lower loss under joint training, together with a guarantee under convexity and additional assumptions; their ablations find that affinities change over the course of training, and that changing the batch size or learning rate can change which tasks should be trained together. That task relationships vary with training state is therefore not new. Our setting differs in four respects: LoRA post-training of a language model rather than vision multi-task networks; fixed per-source exposure and per-expert budgets, so that groupings are constrained on both margins; evaluation by executing each candidate's training interval and scoring its routed experts; and probes measured at a different state from the one at which candidates are trained and deployed. We did not run TAG's procedure, so our results neither replicate nor refute it. In NLP, \citet{vu2020exploring} map intermediate-task transfer among 33 tasks and predict it with task embeddings, finding that transfer can help even when the source task looks dissimilar to the target. For language models, Skill-It \citep{chen2023skillit} estimates a graph of which skills help which others, in its cheaper variant by training on each skill alone and recording whose losses decrease, and uses the graph to order and mix training data for a single model; we use comparable single-source measurements to decide which sources share an expert.

\paragraph{Expert specialization and domain discovery.}
Branch-Train-Merge and Branch-Train-MiX train one expert per given domain and then combine them \citep{li2022btm,sukhbaatar2024btx}. When domain labels are unavailable, c-BTM clusters the corpus by content and trains an expert per cluster \citep{gururangan2023cbtm}; transfer probes are a response-based alternative to such content-based discovery. Classical mixtures of experts and clustered federated learning assign data to models by fit \citep{jacobs1991moe,jordan1994hme,ghosh2020ifca}. DoReMi learns domain mixture weights for training a single model \citep{xie2023doremi}; we instead hold every source's exposure fixed and decide which expert receives it.

\paragraph{Training influence and gradient interaction.}
That training on one source can help or hurt another motivates multi-task learning and gradient-conflict methods \citep{caruana1997multitask,yu2020pcgrad}. TracIn and LESS estimate the effect of training examples on a target through gradient inner products, LESS for instruction-data selection with LoRA \citep{pruthi2020tracin,xia2024less}. Our probes instead measure a source's effect by training on it for a fixed dose and scoring paired outcomes.

\section{Setting}\label{sec:setting}

\paragraph{Sources, experts, and blocks.}
A platform serves requests from $K$ sources---business lines in the motivating setting---and every request carries its source label. The platform maintains $M<K$ experts, each a low-rank adapter $\phi_i$ on one frozen base model \citep{hu2022lora}. Experts are trained in intervals of $H$ source-pure blocks; a block is a run of training examples from one source, with at least a fixed number of completion tokens, applied to one expert. A training allocation is an integer block matrix $n=(n_{ik})$, where $n_{ik}$ counts the blocks of source $k$ that expert $i$ receives in the interval. We hold every expert's budget and every source's exposure fixed, so the feasible allocations are
\begin{equation}
\mathcal N=\Bigl\{n\in\mathbb Z_{\ge 0}^{M\times K}:\ \textstyle\sum_{k}n_{ik}=H/M\ \text{for every } i,\quad \sum_{i}n_{ik}=H/K\ \text{for every } k\Bigr\}.
\label{eq:feasible}
\end{equation}
The row margins $H/M$ fix the expert budgets and the column margins $H/K$ fix the source exposures. Two feasible allocations therefore train on the same tokens from each source and differ only in which expert receives them. Dividing by $H$ gives the transportation polytope $\{Q\ge0:\ Q\mathbf 1_K=\mathbf 1_M/M,\ Q^\top\mathbf 1_M=\mathbf 1_K/K\}$, over which one candidate is computed by linear programming (Section~\ref{sec:probes}).

\paragraph{Serving routes.}
A serving route $h:\{1,\dots,K\}\to\{1,\dots,M\}$ sends every request of source $j$ to expert $h(j)$. We distinguish two ways of fixing it. \emph{Block-matrix routing} reads the route off the allocation: each source is served by the expert that received most of its blocks in the interval, with ties going to the lowest index,
\begin{equation}
h_{\rm blk}(n)(j)=\min\Bigl\{i:\ n_{ij}=\max_{a}n_{aj}\Bigr\}.
\label{eq:blockrouting}
\end{equation}
It depends on $n$ alone. \emph{Competence-calibrated routing} instead scores every trained expert on every source using an independent route-calibration split and serves each source with the best-scoring expert, $h_{\rm cal}(j)\in\arg\max_i\widehat{\mathrm{acc}}^{\rm cal}_{i,j}$; it depends on the trained experts and on additional calibration data. This study uses block-matrix routing only. Competence-calibrated routing was not compared, and the candidate artifacts contain neither route-calibration results nor test predictions from non-routed experts, so it cannot be evaluated after the fact without new generation. Under the uniform allocation every source is a tie, and expert 0 serves all sources.

\paragraph{Utility.}
Let $\Theta_S(n)$ denote the experts obtained by training from state $S$ for one interval with allocation $n$, and let $\mathrm{acc}_{i,j}(\Theta)$ be the probability that a sampled answer of expert $i$ to a source-$j$ problem passes the source-$j$ verifier. For traffic weights $w$ in the probability simplex, the routed utility is
\begin{equation}
U_w(n,h)=\sum_{j=1}^{K}w_j\,\mathrm{acc}_{h(j),j}\bigl(\Theta_S(n)\bigr).
\label{eq:utility}
\end{equation}
All experiments use equal weights, $w_j=1/K$, and the route $h=h_{\rm blk}(n)$; we write $U_{\rm all}$ for this quantity. A candidate is therefore compared through its training allocation together with the route that the allocation implies: both change from one candidate to the next. Because $U_w$ is linear in $w$, the difference between two candidates under any traffic weights lies between their smallest and largest per-source differences; Section~\ref{sec:results} uses this to examine other traffic mixes without new evaluation.

\paragraph{Label reference.}
Suppose the sources come with a label tree. Cutting the tree into $M$ groups and balancing the groups onto the margins of Eq.~\eqref{eq:feasible} yields the allocation that a practitioner holding the tree would choose. We call it the \emph{label reference} C0. The evaluator knows it; a method in the label-unknown setting does not. C0 is a reference point with a stated information source, not an assumed best allocation. The question is whether probes recover a grouping comparable to it, or at least beat an allocation that uses no information.

\paragraph{Testbed.}
Seven sources stand in for business lines: Integral Calculus, Differential Calculus, Abstract Algebra, Linear Algebra, and Statistics from DeepMath-103K \citep{he2025deepmath}, and algorithmic and generic programming problems from OpenCodeInstruct \citep{ahmad2025opencodeinstruct}. Their label tree is mathematics\{calculus\{Integral, Differential\}, algebra\{Abstract, Linear\}, Statistics\} and code\{algorithmic, generic\}. We chose the sources so that the label hierarchy and the response structure could disagree: the tree's four-group cut keeps the two calculus sources together and the two algebra sources together, while Statistics forms a group of its own. Mathematics answers are scored by equivalence with the reference answer; a code answer counts as correct only if it passes every unit-test assertion. Each source's problems are split by hashing into training, feedback, route-calibration, audit, and test sets (66, 12, 6, 10, and 6\%); the probes and the candidate comparison use only the training, audit, and test sets. The test set of algorithmic programming also contains the 376 problems of MBPP+ \citep{liu2023evalplus}, an external benchmark of short Python tasks that carries no algorithmic or generic label; the data manifest assigns all of them to this source as evaluation-only items, scored with the EvalPlus extended tests, and none is used for training or probing. Test problems are taken in sorted identifier order, so the 400 algorithmic test problems used for scoring are these 376 MBPP+ problems and 24 OpenCodeInstruct held-out problems, whereas the audit problems on which this source was probed all come from OpenCodeInstruct. The base model is Qwen3.5-2B-Base \citep{qwen35}, and each expert is a rank-32 LoRA adapter trained with completion-loss supervised fine-tuning. We use $M=4$ experts, $K=7$ sources, and $H=56$ blocks per interval, each with at least 32{,}768 completion tokens (35{,}144 on average in the candidate interval), so each expert receives 14 blocks and each source 8 blocks, and every candidate trains on the same tokens from each source. Answers are sampled at temperature 0.8 with at most 8,192 new tokens for mathematics and 1,024 for code. Besides $U_{\rm all}$ we report $U_{\rm LS}$, the mean accuracy over Linear Algebra and Statistics, and $U_{\rm other5}$, the mean over the other five sources.

With $M=4$ the tree has exactly one cut into four groups, so C0 is fixed by the tree and budget balancing alone (Section~\ref{sec:probes}). The testbed does not favor C0 by construction: in an additive proxy built from development-time probes, the highest-scoring allocation at $M=4$ groups the sources differently from the tree and is predicted to score 0.4426, against 0.4348 for a tree-distance-minimizing fill of the tree grouping and 0.4336 for C0 itself, a predicted margin below the 2 pp minimum useful gain (Section~\ref{sec:probes}).

\paragraph{Scope relative to the motivating platform.}
Table~\ref{tab:scenario} lists which features of a multi-line platform the testbed simulates, which it simplifies, and which it does not cover. The enterprise setting motivates the question only; all data are public, and we do not extrapolate the results to enterprise data.

\begin{table}[t]
\centering\footnotesize
\caption{The testbed against the motivating multi-line platform.}
\label{tab:scenario}
\begin{tabularx}{\linewidth}{@{}>{\raggedright\arraybackslash}p{0.13\linewidth}>{\raggedright\arraybackslash}X>{\raggedright\arraybackslash}X>{\raggedright\arraybackslash}X@{}}
\toprule
Platform aspect & Simulated & Simplified & Not covered \\
\midrule
Lines and labels & Every request carries its line label; seven sources act as lines & Lines are public mathematics subfields and programming types with automatic verifiers & Enterprise data; noisy or changing labels; lines without automatic scoring \\
Expert budget & Fewer experts than lines ($M=4<K=7$); LoRA experts on one shared base & Equal expert budgets and equal line exposure in every interval (Eq.~\ref{eq:feasible}) & Choosing $M$; unequal data volumes, budgets, or hosting costs \\
Knowledge of the grouping & The default and probe-derived candidates use no labels; the tree's grouping is held by the evaluator as the label reference & A tree exists, and at $M=4$ it has a single four-group cut & Platforms with no tree at all, which motivate the study but are not tested \\
Traffic & Per-line traffic weights in the utility (Eq.~\ref{eq:utility}) & Equal weights; other static mixes examined after the fact through linearity (Table~\ref{tab:weightfree}) & Drifting traffic; new or retired lines \\
Training & One allocation interval from a shared warm-up state W & One starting state; one training run per candidate & Repeated reallocation over many intervals; training-seed variation; later states and longer budgets \\
Serving & Each line served by a single expert chosen by a fixed rule & Route read off the block matrix (Eq.~\ref{eq:blockrouting}) & Competence-calibrated routing; latency and throughput of multi-adapter serving \\
Evidence for grouping & Single-source transfer probes with paired scoring & Probes at the initial state only; 16-block dose; 80 audit problems per line; two training-stream shuffles & Probes at the deployment state; probe cost at platform scale \\
Evaluation & Verifier-scored accuracy per line under sampled decoding & 400 test problems per line, two samples each; for algorithmic programming mostly MBPP+, a different benchmark from that line's training and audit problems & User-facing quality measures; greedy or production decoding \\
\bottomrule
\end{tabularx}
\end{table}

\section{Probes and candidate construction}\label{sec:probes}

\paragraph{Probe protocol.}
A transfer probe starts from the initial state $S_0$, the base model with a freshly initialized adapter ($B=0$). It trains the adapter on one source for 16 blocks and scores the 80 audit problems of every source before and after training; adapter and optimizer state are then restored before the next source. A probe row thus gives, for every scored source $j$, the paired accuracy change $D_{kj}$ after training on source $k$, together with the numbers of problems that turned correct ($+$) and incorrect ($-$). We ran two training-stream shuffles: s17, with all seven rows, and s23, with six (the generic-programming row is missing). The shuffles change only the order in which training blocks are drawn. They share the audit problems and the initial predictions, so they are not independent problem-level replications. Each cell is tested with a two-sided exact McNemar test, i.e., a binomial test on the discordant problems. Appendix~\ref{app:stats} gives all cell counts and the multiplicity analysis; here we summarize the statistics that shaped the candidates.

\paragraph{Individual cells were unstable.}
In s17, five of the 49 cells reach nominal $p<0.05$: Abstract$\to$Abstract ($p=0.043$), Linear$\to$Integral ($0.006$), Linear$\to$Statistics ($0.004$; $+16/-3$), Statistics$\to$Linear ($0.016$; $+22/-8$), and Statistics$\to$Statistics ($0.004$); Integral$\to$Abstract ($p=0.0501$) is not among them. In s23, Linear$\to$Statistics recurs at the nominal level ($+14/-4$, $p=0.031$), Statistics$\to$Linear narrowly misses it ($+19/-8$, $p=0.052$), and Differential$\to$Statistics is $+20/-1$. Under a Bonferroni correction over the 49 cells, no s17 cell passes; in s23, Differential$\to$Statistics passes ($p_{\rm Bonf}=0.00103$) and is the only cell that does; no edge passes in both shuffles.

\paragraph{One column was consistent.}
With the four other mathematics sources as donors, pooled over the eight source--shuffle measurements, Statistics gains $+111/-36$; the strongest donor changes between shuffles. This pooled count is descriptive, since the measurements share problems and their exchangeability is not verified. Generic programming has the same baseline accuracy as Statistics in this panel (0.2000) but shows $+55/-60$ ($p\approx0.71$) from the same donors, so the Statistics column is not explained by baseline accuracy alone in this panel. The label tree, by contrast, gives Statistics an expert of its own. The Linear--Statistics pair, pooled over both directions and both shuffles, is $+71/-23$ (descriptive). We singled this pair out after viewing the full table; it is not stronger than the code-domain label pair (algorithmic$\leftrightarrow$generic, $+44/-22$; Fisher exact $p=0.284$ for the difference); and an independent confirmation reserved for it was never run.

\paragraph{Candidates.}
Table~\ref{tab:candidates} lists the three candidates of the main comparison, all fixed from development information before any candidate was trained.
\emph{C0, the label reference,} takes the only four-group cut of the tree and balances it onto the margins.
\emph{C1, the uninformed default,} gives every expert two blocks of every source and uses no information about the sources; it repeats the allocation of the warm-up interval that produced W.
\emph{C2, the probe-derived candidate,} uses the s17 probe table and the margins, and nothing from the tree. The development-time proxy predicts expert $i$'s accuracy on source $j$ as
\[
\hat c_{i,j}(n)=\operatorname{clip}_{[0,1]}\Bigl(\beta_j+\sum_{k}D_{kj}\,\frac{\min(n_{ik},16)}{16}\Bigr),
\]
where $\beta_j$ is the audit accuracy at $S_0$, and scores an allocation by $\frac1K\sum_j\max_i\hat c_{i,j}(n)$, that is, with competence routing inside the proxy. We maximized this score by enumerating serving routes and, for each route, solving a transportation linear program over the continuous relaxation of Eq.~\eqref{eq:feasible}. The optimal vertex, scaled to 56 blocks, is already integral, so integer rounding leaves it unchanged; the proxy scores it at 0.4426. C2 places Linear Algebra with Statistics (expert e0: 6 and 8 blocks), keeps the two code sources together (e2), pairs Integral Calculus with Abstract Algebra (e3), and gives Differential Calculus an expert (e1) that also takes two blocks each of Abstract Algebra, Linear Algebra, and generic programming. Under block-matrix routing e0 serves both Linear Algebra and Statistics, so C2 is the main comparison's label-free test of the Linear--Statistics grouping that the probes suggested.

The same run also executed a fourth candidate, C3, that edits C0 by exchanging Abstract Algebra and Statistics between two experts. Two of its groups, the calculus and code experts, are copied verbatim from the label reference, and its Abstract Algebra expert is C0's Statistics expert with Abstract Algebra in place of Statistics, so it cannot be constructed without the tree and lies outside the label-unknown setting. C3 was part of the frozen protocol; classifying it as outside the label-unknown setting was decided after observing results, when the report was reframed around that setting. Appendix~\ref{app:c3} reports it in full.

\begin{table}[t]
\centering\small
\caption{Candidates in the main comparison. Block matrices list experts e0--e3 as rows and sources (Integral, Differential, Abstract, Linear, Statistics, algorithmic, generic) as columns; every row sums to 14 and every column to 8. Routes are block-matrix routes (Eq.~\ref{eq:blockrouting}) in the same source order.}
\label{tab:candidates}
\begin{tabular}{@{}lccp{0.38\linewidth}@{}}
\toprule
Candidate & Blocks $n$ & Route $h_{\rm blk}(n)$ & Role and information used \\
\midrule
C0 & $\left(\begin{smallmatrix}0&0&7&7&0&0&0\\7&7&0&0&0&0&0\\0&0&0&0&0&7&7\\1&1&1&1&8&1&1\end{smallmatrix}\right)$ & $(1,1,0,0,3,2,2)$ & Label reference: the tree's only four-group cut, balanced onto the margins. Uses the tree; not available to a method in the label-unknown setting. \\[3pt]
C1 & $\left(\begin{smallmatrix}2&2&2&2&2&2&2\\2&2&2&2&2&2&2\\2&2&2&2&2&2&2\\2&2&2&2&2&2&2\end{smallmatrix}\right)$ & $(0,0,0,0,0,0,0)$ & Uninformed default: uses no information. Every source is a tie, so expert 0 serves all sources. \\[3pt]
C2 & $\left(\begin{smallmatrix}0&0&0&6&8&0&0\\0&8&2&2&0&0&2\\0&0&0&0&0&8&6\\8&0&6&0&0&0&0\end{smallmatrix}\right)$ & $(3,1,3,0,0,2,2)$ & Probe-derived: transportation-LP vertex for the additive proxy built from the s17 probe table. Uses the probes and the margins, not the tree. \\
\bottomrule
\end{tabular}
\end{table}

\paragraph{Uniqueness of C0.}
The tree's group sizes are two, two, one, and two sources, which equal expert budgets cannot hold as a hard partition: 7 sources of 8 blocks give 56 blocks, and the groups would need 16, 16, 8, and 16. Balancing therefore moves 6 of the 56 blocks out of their preferred expert. Projecting the hard source-to-expert map onto Eq.~\eqref{eq:feasible} by Sinkhorn matrix scaling \citep{sinkhorn1967} and rounding gives the calculus, algebra, and code experts 7 blocks of each of their two sources, and the Statistics expert 8 Statistics blocks plus one block of every other source. This fill depends only on the labels and the margins; it is also one of the tied minimizers of a tree-distance transportation cost. The tree has a single four-group cut at $M=4$ (for $M=2,\dots,7$ the numbers of cuts are 1, 1, 1, 3, 3, and 1), so any rule that selects a cut, including one that scores cuts on development data, has only one option at $M=4$, and no development response data influenced C0. The setting was not tilted toward C0 either: under the development-time proxy above, the highest-scoring allocation groups the sources differently from the tree (0.4426 against 0.4348 for a tree-distance-minimizing fill of the tree grouping, both under competence routing inside the proxy; C0 itself scores 0.4336). This predicted advantage, $+0.78$\,pp over that fill and $+0.90$\,pp over C0, is below the minimum useful gain $\delta_{\min}=2$\,pp of the comparison (Section~\ref{sec:results}): the margin the probes predicted at development time was smaller than the gain the comparison was designed to detect. Evaluated after the fact under the executed block-matrix routing, the same proxy places C2 $+2.01$\,pp (s17) and $+0.67$\,pp (s23*) above C0 (Appendix~\ref{app:proxy}).

\paragraph{From probe to score.}
The probes, the candidates, and the scores come from different states and different data (Table~\ref{tab:state-evidence}). Between the probes and the comparison, five things changed: the training state ($S_0$ to W), the item bank (80 audit to 400 test problems per source; for algorithmic programming also from OpenCodeInstruct to mostly MBPP+ problems), the dose (16 blocks of one source in a probe versus at most 8 blocks of a source per expert in an interval), the training mix (single-source probes versus mixed-source expert training), and the route (C2 was optimized under competence routing inside the proxy and executed under block-matrix routing).

\begin{table}[t]
\centering\footnotesize
\caption{State and evidence along the chain from probes to scores.}
\label{tab:state-evidence}
\begin{tabularx}{\linewidth}{@{}>{\raggedright\arraybackslash}p{0.13\linewidth}>{\raggedright\arraybackslash}p{0.17\linewidth}>{\raggedright\arraybackslash}X>{\raggedright\arraybackslash}X@{}}
\toprule
Link & State & Data & What it answers, and what it does not \\
\midrule
1.~Probes & $S_0$: base model with a fresh adapter ($B=0$) & Training split, one source, 16 blocks; audit split, 80 problems per source (OpenCodeInstruct for both code sources); shuffles s17 (7/7 rows) and s23 (6/7 rows), sharing problems and initial predictions & Whether training on source $k$ moves accuracy on source $j$ at $S_0$, as paired flips. Not: effects at W, under mixed-source training, or at other doses. \\
2.~Candidate construction (development) & No training; uses the $S_0$ probe table & C0: label tree and margins. C1: margins only. C2: s17 table, margins, and the additive proxy with competence routing. No test data and no data from W & Which allocations to train. Not: how they perform. C3, an edit of C0, is described in Appendix~\ref{app:c3}. \\
3.~Training at W & W: shared warm-up, one uniform interval (two blocks per expert per source) & Training split; one interval of 56 blocks per candidate, with identical tokens per source across candidates; one run per candidate & What each allocation produces from W in one interval. Not: training-seed variation, other starting states, or longer budgets. \\
4.~Fixed-routing test & W plus one candidate interval & Test split, first 400 problems per source (algorithmic: 376 MBPP+ and 24 OpenCodeInstruct problems), two samples each (seeds 2001 and 2002); route by Eq.~\ref{eq:blockrouting} & $U_{\rm all}$, $U_{\rm LS}$, $U_{\rm other5}$, and per-source accuracy of allocation and route together, with intervals for test sampling. Not: performance under competence-calibrated routing; for algorithmic programming, performance on the benchmark that was probed. \\
\bottomrule
\end{tabularx}
\end{table}

\section{Candidate comparison at W}\label{sec:results}

\paragraph{Evaluation protocol.}
The candidates (the three above and the candidate of Appendix~\ref{app:c3}), the routing rule, the test problems, the number of samples, the evaluation seeds, the minimum useful gain, the decision table, and the analysis were frozen before the candidates were trained. Each candidate was trained once from W for one interval (about 4,570\,s) and scored on the first 400 test problems of every source in a fixed sorted order (the smallest test bank has 421; for algorithmic programming the 400 are 376 MBPP+ and 24 OpenCodeInstruct problems), with the same problems for all candidates. Each problem was answered $k=2$ times, with CUDA sampling seeds 2001 and 2002 shared across candidates. The minimum useful gain, $\delta_{\min}=2$\,pp in $U_{\rm all}$, was set from measurement capacity before any candidate result: with one sample per problem and 400 problems per source, the paired standard error of $\Delta U_{\rm all}$ was estimated at 0.98\,pp and the minimum detectable effect (MDE, 80\% power) at 2.74\,pp. A rule fixed in advance chose the smallest $k\in\{1,2,4\}$ with MDE at most $\delta_{\min}$, using a split of decoding variance: two samples from the same model disagreed on 21--34\% of problems per source, decoding accounted for 42--67\% of single-sample variance, and the projected MDE was 2.74, 1.93, and 1.36\,pp for $k=1$, 2, and 4. These design inputs come from an earlier comparison that is not part of this report: the split of decoding variance used one four-expert portfolio trained from W, and the standard error and MDE used the per-item paired difference between two four-expert portfolios of that comparison, both trained from W; only their summary is included in the evidence package (\path{results/protocol/}; Appendix~\ref{app:protocol}). In the candidate round itself, a candidate's two samples disagreed on 20--34\% of problems per source. Intervals come from a joint paired bootstrap with 2000 resamples, in which problems are resampled within each source with the same indices for all candidates, and bounds are percentiles. They cover test sampling (problems and decoding) conditional on each candidate's single training run; they do not include training randomness. Appendix~\ref{app:protocol} gives the details.

\begin{table}[t]
\centering\small
\caption{Candidate comparison at W. Top: routed utilities (mean accuracy; 400 test problems per source, two samples each). Bottom: paired differences in percentage points with 95\% joint paired bootstrap intervals. $U_{\rm LS}$ averages Linear Algebra and Statistics; $U_{\rm other5}$ averages the other five sources.}
\label{tab:results}
\setlength{\tabcolsep}{4pt}
\begin{tabularx}{\linewidth}{@{}l>{\raggedright\arraybackslash}p{0.15\linewidth}>{\centering\arraybackslash}X>{\centering\arraybackslash}X>{\centering\arraybackslash}X@{}}
\toprule
 & When specified & $U_{\rm all}$ & $U_{\rm LS}$ & $U_{\rm other5}$ \\
\midrule
C0 (label reference) & --- & 0.5036 & 0.5019 & 0.5042 \\
C1 (uninformed default) & --- & 0.4782 & 0.4888 & 0.4740 \\
C2 (probe-derived) & --- & 0.4793 & 0.4906 & 0.4747 \\
\midrule
 & & $\Delta U_{\rm all}$ & $\Delta U_{\rm LS}$ & $\Delta U_{\rm other5}$ \\
\midrule
C1 $-$ C0 & before the run & $-2.54$ $[-4.04,-1.12]$ & $-1.31$ $[-4.19,+1.50]$ & $-3.02$ $[-4.70,-1.42]$ \\
C2 $-$ C0 & before the run & $-2.43$ $[-3.84,-1.14]$ & $-1.13$ $[-3.88,+1.38]$ & $-2.95$ $[-4.50,-1.40]$ \\
C2 $-$ C1 & after observing results & $+0.11$ $[-1.25,+1.54]$ & $+0.19$ $[-2.44,+2.81]$ & $+0.07$ $[-1.60,+1.75]$ \\
\bottomrule
\end{tabularx}
\par\smallskip
{\footnotesize\raggedright ``Before the run'': specified before the run in the frozen protocol. ``After observing results'': added after observing results; computed on the same scores with the same bootstrap scheme and a separate bootstrap seed. A fourth executed candidate, which edits the label reference and is therefore outside the label-unknown setting, is reported in Appendix~\ref{app:c3} for completeness.\par}
\end{table}

\paragraph{Reading the comparisons specified before the run.}
The frozen decision table had three rows: a lower bound above $+\delta_{\min}$ indicates an opportunity worth pursuing; an interval entirely inside $(-\delta_{\min},+\delta_{\min})$ indicates that the candidate is unlikely to deliver the specified gain at this state; an interval crossing $\pm\delta_{\min}$ indicates insufficient precision. The analysis script also reports a fourth reading, an interval entirely below $-\delta_{\min}$; it was not part of the frozen table, and no interval reached it. Read literally, both intervals in Table~\ref{tab:results} that were specified before the run contain $-\delta_{\min}$ and fall in the last row of the frozen table. That table was written to detect gains and has no row for a shortfall, however, and these intervals cross $-\delta_{\min}$, not $+\delta_{\min}$. We therefore answer three questions separately. Is a gain of $\delta_{\min}$ supported? No: the upper bounds ($-1.12$ and $-1.14$\,pp) are below $+2$\,pp, and below zero. Did the candidates improve on C0 or fall short of it? Both intervals lie below zero, which supports a shortfall. Does the shortfall reach $\delta_{\min}$? This is not resolved: the C1 and C2 point estimates are near $-2.5$\,pp, but their intervals reach $-1.1$\,pp. The qualified conclusion is:
\begin{quote}
Under the W starting state, a single fixed-budget interval, and the specified block-matrix routing, neither candidate showed the prespecified practical gain, and each had lower overall utility than C0.
\end{quote}
It concerns these two candidates, one starting state, one budget, and one routing rule. It does not make C0 the best of the feasible allocations, and it does not extend to training randomness, since each candidate was trained once.

\paragraph{Probe-derived grouping versus the uninformed default (added after observing results).}
The comparison that matters most for the label-unknown setting, C2 against C1, was not specified in the frozen protocol. On the same scores, C2$-$C1 is $+0.11$\,pp, $[-1.25,+1.54]$, an interval entirely inside $(-\delta_{\min},+\delta_{\min})$; across 22 bootstrap seeds its lower bound ranged from $-1.34$ to $-1.20$ and its upper bound from $+1.43$ to $+1.57$ (Appendix~\ref{app:stats}). The Linear--Statistics grouping that C2 takes from the probes showed no resolved effect on those two sources either: $\Delta U_{\rm LS}$ is $-1.13$, $[-3.88,+1.38]$, against C0 and $+0.19$, $[-2.44,+2.81]$, against C1. Against C0, the shortfall of C2 lies mainly in the other five sources ($\Delta U_{\rm other5}=-2.95$, $[-4.50,-1.40]$), as does that of C1 ($-3.02$, $[-4.70,-1.42]$). At this resolution, the purely probe-derived grouping was indistinguishable from the uninformed default, and both scored below the label reference.

\paragraph{Other traffic mixes and single sources (added after observing results).}
Equal weights are one traffic mix among many. Because $\Delta U_w=\sum_jw_j\Delta_j$ is linear in $w$, its range over all weights is spanned by the per-source differences $\Delta_j$ (Table~\ref{tab:weightfree}): $[-5.875,+0.875]$\,pp for C1 and $[-5.125,+0.875]$\,pp for C2. No traffic mix therefore gives either candidate a point-estimate advantage over C0 above 0.875\,pp, which is below $\delta_{\min}$. This statement concerns point estimates. Single-source differences are imprecise: simultaneous 95\% intervals over the family of all 21 executed per-source comparisons (three non-reference candidates, including C3, times seven sources) need a critical value of 3.025 instead of 1.96 (2.965 to 3.094 across 22 bootstrap seeds), and their upper bounds reach $+6.62$\,pp for C1 (Abstract Algebra) and $+6.45$\,pp for C2 (Statistics). If traffic were concentrated on a single source, a gain of several percentage points could not be excluded. Only C1 on generic programming is entirely below zero for every bootstrap seed checked. C2 on algorithmic programming is borderline, and in both C0 and C2 that source is served by an expert trained only on the two code sources (7+7 and 8+6 blocks), so its difference cannot be attributed to a change of grouping; its test problems are, moreover, mostly MBPP+ problems, a different benchmark from the OpenCodeInstruct problems on which the source was trained and probed. Single-source point differences reached about 3.6\,pp on control sources whose training allocation and routing were matched between C0 and C3 (Appendix~\ref{app:protocol}); these controls showed no evident anomaly, but they caution against reading single-source differences one at a time.

\begin{table}[t]
\centering\small
\caption{Other traffic mixes and single sources, relative to the label reference (percentage points). $\Delta_j$ is the per-source accuracy difference. Because utility is linear in the traffic weights, $\Delta U_w=\sum_jw_j\Delta_j$ lies in $[\min_j\Delta_j,\max_j\Delta_j]$ for every $w$. Intervals are simultaneous 95\% intervals (max-$|t|$ bootstrap, critical value 3.025).}
\label{tab:weightfree}
\begin{tabular}{@{}lrcrc@{}}
\toprule
 & \multicolumn{2}{c}{C1 $-$ C0} & \multicolumn{2}{c}{C2 $-$ C0} \\
\cmidrule(lr){2-3}\cmidrule(l){4-5}
Source & $\Delta_j$ & Simultaneous 95\% & $\Delta_j$ & Simultaneous 95\% \\
\midrule
Integral Calculus & $-4.500$ & $[-10.42,+1.42]$ & $-3.250$ & $[-8.87,+2.37]$ \\
Differential Calculus & $-4.250$ & $[-10.02,+1.52]$ & $-4.625$ & $[-10.14,+0.89]$ \\
Abstract Algebra & $+0.875$ & $[-4.87,+6.62]$ & $-0.125$ & $[-5.73,+5.48]$ \\
Linear Algebra & $-1.250$ & $[-7.57,+5.07]$ & $-3.125$ & $[-8.85,+2.60]$ \\
Statistics & $-1.375$ & $[-7.10,+4.35]$ & $+0.875$ & $[-4.70,+6.45]$ \\
algorithmic programming & $-1.375$ & $[-6.99,+4.24]$ & $-5.125$ & $[-10.18,-0.07]^{\dagger}$ \\
generic programming & $-5.875$ & $[-11.21,-0.54]^{*}$ & $-1.625$ & $[-6.39,+3.14]$ \\
\midrule
Range of $\Delta U_w$ over all $w$ & \multicolumn{2}{c}{$[-5.875,+0.875]$} & \multicolumn{2}{c}{$[-5.125,+0.875]$} \\
Equal weights ($\Delta U_{\rm all}$) & \multicolumn{2}{c}{$-2.54$} & \multicolumn{2}{c}{$-2.43$} \\
\bottomrule
\end{tabular}
\par\smallskip
{\footnotesize\raggedright The per-source point differences were reported by the analysis specified before the run; the ranges over traffic weights and the simultaneous intervals were added after observing results. Every $\Delta_j$ is a multiple of 0.125\,pp and is printed exactly. The simultaneous family has 21 members and includes the seven differences of C3 (Appendix~\ref{app:c3}), so the intervals are conservative for the 14 shown. $^{*}$Entirely below zero; the upper bound stays below zero for all 22 bootstrap seeds checked ($-0.82$ to $-0.29$). $^{\dagger}$Borderline: the upper bound is $-0.07$ at the post hoc bootstrap seed (20261009) and $+0.03$ when the post hoc simultaneous interval is recomputed with the bootstrap seed of the primary analysis (20261007), and ranges from $-0.07$ to $+0.23$ across 22 seeds (below zero for 7 of them); we do not treat it as resolved.\par}
\end{table}

\paragraph{Proxy diagnostic (added after observing results).}
After observing the results we compared the additive proxy that produced C2 with the measurements (Appendix~\ref{app:proxy}). Under block-matrix routing, the proxy built from the s17 table predicted C1$-$C0 $=-2.23$\,pp and C2$-$C0 $=+2.01$\,pp; with the s23 table, whose missing generic row is borrowed from s17, it predicted $-1.92$ and $+0.67$\,pp; with competence routing inside the proxy (s17), $-3.90$ and $+0.90$\,pp. The measured differences were $-2.54$ and $-2.43$\,pp. Keeping only the diagonal of the table, i.e., assuming no transfer between sources, the proxy predicted $-3.73$ and $+0.02$\,pp (s17). Within this proxy, then, the explanation that transfer vanished at W is not sufficient to explain why C2 scored below C0: without transfer, the proxy places C2 level with C0, not below it. The diagnostic cannot show that the real shortfall comes from interference in joint training. Of the five changes listed in Section~\ref{sec:probes}, the readings above vary the routing and the proxy has no term for the training mix; for the other three, proxy and measurement differ in state ($S_0$ versus W), item bank (80 audit versus 400 test problems per source, and for algorithmic programming a different benchmark), and dose (16 blocks of one source versus at most 8 per source and expert); C2 is the optimum of this same proxy, so its proxy prediction is biased upward (the optimizer's curse); and there are only three executed non-reference candidates, two of them in the main comparison. Routing alone moves the proxy by an amount comparable to these predictions: switching C0 from block-matrix to competence routing raises its predicted utility by 1.66\,pp (s17; 1.75\,pp with s23).

\section{Interpretation boundaries and lessons}\label{sec:lessons}

This section separates what the comparison resolves, which expectations it did not bear out, and which explanations it cannot tell apart, and then lists lessons that this run supports. Everything here is conditional on one starting state (W), one allocation interval, block-matrix routing, and one training run per candidate.

\paragraph{What is resolved.}
Under the W starting state, a single fixed-budget interval, and the specified block-matrix routing, neither the uninformed default C1 nor the probe-derived grouping C2 showed the prespecified practical gain, and each had lower overall utility than the label reference C0 (Table~\ref{tab:results}); the additional executed candidate reported in Appendix~\ref{app:c3} also scored below C0. Three questions have different answers. A gain of 2 pp is not supported: both upper limits ($-1.12$ and $-1.14$ pp) are below $+2$ pp, and below zero. A shortfall relative to C0 is supported. Whether the shortfall reaches 2 pp is not resolved. The statement concerns the combination of a training allocation and the routing computed from it. It does not make C0 the best feasible allocation, and it does not extend to training randomness, because the intervals cover only test items and decoding.

\paragraph{Expectations that were not borne out.}
\begin{itemize}
\item \emph{The probe-derived grouping would score above the label reference.} A response model fitted to development probes predicted that a grouping different from the tree's would score higher (Section~\ref{sec:setting}), and C2 is the allocation that maximizes that model; at development time, under the model's own competence routing, its predicted advantage over C0 was $+0.90$ pp, below $\delta_{\min}$ (Section~\ref{sec:probes}). Evaluated afterwards with the executed block-matrix routing, the same additive model placed C2 $+2.01$ pp above C0 with the first shuffle's tensor and $+0.67$ pp with the second (Appendix~\ref{app:proxy}). Measured, C2 was $2.43$ pp below C0, $[-3.84,-1.14]$. Because C2 was chosen by maximizing the model, its predicted advantage is inflated by that selection; the model is a diagnostic here, not an independent forecast.
\item \emph{Probe information would beat no information.} In a comparison added after observing results, C2 $-$ C1 was $+0.11$ pp, $[-1.25,+1.54]$. The interval lies inside $\pm2$ pp, so a difference of the prespecified practical size is not supported in either direction.
\item \emph{Grouping Linear Algebra with Statistics would help those two lines.} The probes suggested this pairing (Linear$\to$Statistics $+16/-3$ and Statistics$\to$Linear $+22/-8$ in the first shuffle), and C2 contains it: its expert e0 trains on 6 Linear Algebra and 8 Statistics blocks and serves both lines, so the pairing was executed inside the label-unknown setting. C2's $\Delta U_{\rm LS}$ was $-1.13$ pp $[-3.88,+1.38]$ against C0 and, post hoc, $+0.19$ pp $[-2.44,+2.81]$ against C1, whereas the additive model had predicted $+4.92$ and $+1.95$ pp against C0 for the two shuffles. The evidence for the pairing was nominal only: it was singled out after inspecting the $7\times7$ tensor, none of its cells passed a 49-cell Bonferroni correction, and the confirmation run reserved for it was never executed (Appendix~\ref{app:stats}).
\end{itemize}

\paragraph{Explanations this run cannot separate.}
Between the probes and the comparison several things changed at once (Table~\ref{tab:state-evidence}), and the design isolates none of them.
\begin{itemize}
\item \emph{Routing.} Each candidate was served by the routing computed from its own block matrix (Eq.~\eqref{eq:blockrouting}), so training allocation and serving routing changed together. Under C1 every source is a tie, and the tie-break gives all seven lines to expert e0. Competence-calibrated routing, which chooses experts on an independent calibration bank, was not compared, and the candidate artifacts contain neither route-calibration results nor test predictions from non-routed experts, so that comparison cannot be reconstructed without new generation. In the additive model, switching C0 alone from block-matrix to competence routing raised its predicted utility by $1.66$ pp (first shuffle) and $1.75$ pp (second), the same order as the gaps at issue (Appendix~\ref{app:proxy}); these are model figures, not measurements.
\item \emph{Joint training.} A probe trains one source on a fresh adapter. A candidate expert trains on two to seven sources within one interval; C2's expert e0, for instance, receives Linear Algebra and Statistics blocks together. Interference among co-trained sources could offset transfer measured in isolation, and the additive model has no term for it. Removing every transfer term from that model brings C2 $-$ C0 to $+0.02$ pp (first shuffle) and $-0.10$ pp (second), not to the measured $-2.43$ pp, so within the model the loss of probed transfer by state W does not by itself account for the shortfall. This does not show that the shortfall comes from joint-training interference: the model's state, item bank, and dose all differ from the comparison's, and it has been checked against only a few executed points.
\item \emph{Training randomness.} Each candidate was trained once, and every interval covers test-item and decoding sampling only. For the two experts that received identical blocks in C0 and in the candidate of Appendix~\ref{app:c3}, final parameters differed by about 2\% of their training displacement from GPU nondeterminism alone; data order and initialization were not varied, and no candidate was retrained.
\item \emph{State drift, item bank, and dose.} Probes were measured at $S_0$ (the base model with a freshly initialized adapter), on the audit split (80 items per source), after 16 blocks of a single source. Candidates started from W, reached after one uniform warm-up interval, were scored on the test split (400 items per source; for algorithmic programming mostly MBPP+ items), and split each source's 8 blocks of the interval among the experts (2 per expert under C1). Any of these differences could change the size or the sign of a probed effect, and the run contains no measurement that separates them. Section~\ref{sec:open} describes measurements that would separate state drift, routing, training randomness, and joint training (items~\ref{oq:state}--\ref{oq:interference}); none of them was run.
\end{itemize}

\paragraph{What remains open.}
Traffic reweighting does not rescue either candidate on point estimates: because utility is linear in traffic weights, any reweighting yields a difference from C0 between the worst and the best per-line difference, and the best per-line difference is $+0.875$ pp for both C1 (Abstract Algebra) and C2 (Statistics), below $\delta_{\min}=2$ pp (Table~\ref{tab:weightfree}). Per-line resolution is poor, however: simultaneous upper limits reach $+6.62$ pp (C1, Abstract Algebra) and $+6.45$ pp (C2, Statistics), so a large gain for traffic concentrated on a single line is not excluded. Whether probes measured at the deployment state or candidates served with competence-calibrated routing would change the comparison is not addressed by this run; Section~\ref{sec:open} lists these and other open questions with the data and compute they would need.

\paragraph{Operational lessons.}
Each lesson is backed by a specific observation in this project.
\begin{enumerate}
\item \emph{Check the probe state against the deployment state.} The probes were measured at $S_0$ and acted upon at W, with a different item bank and dose; Table~\ref{tab:state-evidence} makes these mismatches explicit. Record for each measurement the adapter state, item bank, and dose; where the probe state and the decision state differ, re-probe at the decision state or treat transfer there as an open hypothesis.
\item \emph{Register training allocation and serving routing separately.} In this round both changed with the candidate, so a shortfall cannot be assigned to either; under C1 a tie-break rule, not a measurement, decided that one expert served every line. Fix routing across arms, or evaluate each trained portfolio under more than one routing rule and keep the predictions of non-routed experts, so that routing can be re-analyzed without new training.
\item \emph{Size evaluations for sampled decoding.} Two samples from the same model disagreed on 20--34\% of test items per source, and in the design measurement decoding accounted for 42--67\% of single-sample variance (Appendix~\ref{app:protocol}). With 400 items per source and two samples per item, the 95\% interval for an overall difference had a half-width of about 1.4 pp, whereas the 21 executed per-line differences (including those of the appendix candidate) required a simultaneous critical value of 3.03 instead of 1.96 and had simultaneous half-widths between about 4.5 and 6.3 pp (Appendix~\ref{app:stats}). Line-level conclusions need many more items than conclusions about overall utility.
\item \emph{State the information source of a label reference.} Describing C0 as the tree cut that scores best on development response data would suggest that the reference had used response data. With four experts the tree has exactly one four-group cut, so such a selection is vacuous and C0 is fixed by the tree and budget balancing alone (Section~\ref{sec:probes}). What a comparison against a reference shows depends on what the reference was allowed to see.
\end{enumerate}

\section{Open questions}\label{sec:open}

This run leaves the questions below open. For each we state what it would distinguish, what data it needs, and a rough cost. Costs are extrapolated from the ledger of the candidate round (\path{results/candidates/training_summary.json} in the evidence package), with every process on a single GPU: training one 56-block interval took $t_{\rm train}\approx4{,}570$\,s ($\approx1.3$ GPU-hours, about 82\,s per block), and one evaluation of 400 items per source with one sample each, $7\times400=2{,}800$ generations, took $t_{\rm eval}\approx32{,}300$\,s ($\approx9.0$ GPU-hours, $g\approx32{,}300/2{,}800\approx11.5$\,s per generation). One candidate as executed here, one interval and $k=2$ evaluations, therefore costs $t_{\rm train}+2t_{\rm eval}\approx69{,}200$\,s $\approx19.2$ GPU-hours. A full probe tensor at the design of Section~\ref{sec:probes} trains 7 rows of 16 blocks and scores $7\times80=560$ audit items once before training and once after each row, so it costs about $112\times82\,{\rm s}+8\times560\times11.5\,{\rm s}\approx61{,}000$\,s $\approx17$ GPU-hours per shuffle. Generation time varies with the source mix, the generation limits, and the model state, and larger base models raise both training and generation costs, so these figures give orders of magnitude only.
\begin{enumerate}
\item\label{oq:state} \emph{Do probes at the decision state predict grouping value?} This would distinguish state drift ($S_0$ versus W) from the other changes between probe and comparison: if probes measured at W still favor a Linear--Statistics grouping and a candidate built from them still trails C0, state drift alone does not explain the shortfall. Data: a transfer tensor measured from one of W's expert adapters with the same dose and audit items, in two shuffles, and one candidate built from it by the same transportation linear program, trained and scored against C0 and C1 on the same items and seeds. Cost: $2\times17+19.2\approx53$ GPU-hours; the comparison with the existing C0 and C1 runs also carries their training randomness (item~\ref{oq:seed}).
\item\label{oq:routing} \emph{Does the comparison change under competence-calibrated routing?} This would separate the effect of the allocation from that of the route; under block-matrix routing C1 serves every line from expert 0 by a tie-break, and in the additive model competence routing alone raised C0's predicted utility by 1.66\,pp (Appendix~\ref{app:proxy}). Data: scores of every expert on every source on the route-calibration split, which is independent of the test split and has at least 356 items per source, and test predictions of the experts the calibrated routes select, ideally of all experts so that routing can be re-analyzed later. Cost per trained portfolio, with 80 calibration items per source: $4\times7\times80=2{,}240$ generations ($\approx7.2$ GPU-hours) plus at most one full $k=2$ test evaluation under the new routes ($2t_{\rm eval}\approx18$ GPU-hours), so at most about 25 GPU-hours, or 75 for C0--C2; keeping test predictions of all four experts instead costs $4\times18=72$ GPU-hours per portfolio. The trained checkpoints are not in the evidence package; retraining one costs about 1.3 GPU-hours.
\item\label{oq:seed} \emph{Does the ranking survive training randomness?} Every interval in this report is conditional on one training run per candidate; GPU nondeterminism alone moved matched experts by about 2\% of their training displacement (Appendix~\ref{app:protocol}). Repeated runs would show whether C0's advantage exceeds run-to-run variation. Data: additional training runs of each candidate from W with varied data order and adapter initialization, and separately with identical seeds to isolate nondeterminism, scored on the same items and seeds. Cost: 19.2 GPU-hours per run; one replicate of C0 and C2 costs $2\times19.2\approx38$ GPU-hours, three replicates of C0--C2 cost $9\times19.2\approx173$ GPU-hours; varying the starting state adds a warm-up interval ($\approx1.3$ GPU-hours) and a full set of candidate runs per state.
\item\label{oq:interference} \emph{Can joint-training interference explain the shortfall below the zero-transfer prediction?} In the additive model, removing every transfer term places C2 level with C0 ($+0.02$\,pp), whereas the measurement is $-2.43$\,pp, and the model has no term for sources trained together on one expert. A direct measurement would compare, from W, each multi-source expert of a candidate with single-source experts trained on the same blocks of each of its sources, scored on those sources. Cost for an expert with $s$ sources: $28$ trained blocks ($14$ mixed and $14$ single-source, $28\times82\,{\rm s}\approx0.6$ GPU-hours) and $2s\times800$ generations at $k=2$ ($\approx10.3$ GPU-hours for $s=2$); for C2's four experts, three with two sources and one with four, about $3\times10.9+21.1\approx54$ GPU-hours. Per-source effects near 2\,pp would need the item counts of item~\ref{oq:resolution}.
\item\label{oq:content} \emph{Does a cheap label-free alternative do as well?} Content-based grouping, as in c-BTM \citep{gururangan2023cbtm}, needs neither labels nor probe training. Comparing it with the probe-derived C2 and the uniform C1 would show whether probes add anything over content similarity, and whether content similarity recovers the label tree. Data: embeddings of a sample of each source's training examples, a clustering of the seven source centroids into four groups, and the same balancing and rounding as for C0; then one candidate trained and scored as here. Cost: one forward pass over the sample, small next to a training interval, plus 19.2 GPU-hours for the candidate, or nothing beyond the existing C0 run if the clustering reproduces the tree's four-group cut.
\item\label{oq:general} \emph{Do the findings hold in other testbeds, for other base models, with longer budgets, and with repeated allocation?} This would distinguish a property of this testbed (seven sources, one 2B base model, one interval from W) from a general one, and test whether reallocating between intervals on measured outcomes recovers from a poor initial grouping. Data: other source sets in which label and response structure might disagree, other base models and sizes, several intervals per candidate, and a reallocation rule fixed in advance. Cost per testbed at this model size, with one training run per candidate: two probe shuffles ($\approx34$), a warm-up interval ($\approx1.3$), and three candidates ($3\times19.2\approx58$), about 93 GPU-hours before any replication; each further interval adds $\approx1.3$ GPU-hours of training per candidate and $\approx18$ GPU-hours for every interim $k=2$ evaluation.
\item\label{oq:platform} \emph{Do the lessons carry over to the motivating platform?} Four features of that setting are absent here: requests that need the skills of several lines, which a single line-to-expert route cannot serve; serving without line labels, where a learned router and the grouping interact; traffic that drifts over time, whereas the linearity argument of Section~\ref{sec:results} covers static mixes only; and data isolation, where some lines may not share an expert, which forces zeros in the block matrix and shrinks the feasible set of Eq.~\eqref{eq:feasible}. Data: platform data with these properties, or public-data simulations of them such as composed multi-source requests, an unlabeled request stream, and a traffic sequence. Cost: not estimable from this run; on public data each simulated candidate would cost about the same as here (19.2 GPU-hours per run) plus the router.
\item\label{oq:resolution} \emph{How many items would resolve a single line?} The simultaneous per-line half-widths here were 4.5--6.3\,pp. With $n=400$ items per line and $k=2$, the bootstrap standard errors of per-line differences from C0 were ${\rm se}_{400}=1.48$--$2.09$\,pp (from the simultaneous-interval analysis of Appendix~\ref{app:stats}). Detecting $\delta=2$\,pp on one line with 80\% power at two-sided $\alpha=0.05$, before any multiplicity adjustment, needs about $n\approx400\,(2.8\,{\rm se}_{400}/\delta)^2\approx1{,}700$--$3{,}400$ items per line, more than the test split of every source except generic programming holds (421--1{,}144 items; 5{,}495 for generic programming), so the test pool would have to be enlarged from other splits or new data. Cost: evaluating one candidate at $k=2$ on $n$ items per line takes $14n$ generations, $14n\times11.5$\,s, i.e., about 76--152 GPU-hours per candidate for $n=1{,}700$--$3{,}400$, against 18 GPU-hours for the 400-item evaluation used here. Raising $k$ beyond 2 reduces only the decoding part of the variance.
\end{enumerate}

\paragraph{A minimal credible study.}
Items~\ref{oq:state}, \ref{oq:routing}, \ref{oq:seed}, and~\ref{oq:content} could be addressed together in one testbed at this model size: (i) probes at W in two shuffles, $2\times17\approx34$ GPU-hours; (ii) five candidates (C0, C1, the $S_0$-probe candidate C2, a W-probe candidate, and a content-cluster candidate), each trained three times from W and scored at $k=2$ on 400 items per line under block-matrix routing, $5\times3\times19.2\approx288$ GPU-hours; (iii) competence-calibrated routing for one training run of each candidate, at most $5\times(7.2+18)\approx126$ GPU-hours. The total, $34+288+126\approx450$ GPU-hours, is on the order of 500 GPU-hours. Such a study would separate state drift, routing, and training randomness at the overall-utility resolution of this run (projected MDE 1.93\,pp at $k=2$, without training variance), but it would not resolve single lines (item~\ref{oq:resolution}) or establish generality (items~\ref{oq:general} and~\ref{oq:platform}).

\section{Conclusion}\label{sec:conclusion}

We asked whether transfer probes can discover which lines should share an expert when no trusted taxonomy says so, and tested the idea where a label tree does exist so that its grouping could serve as a label reference. With seven public sources standing in for business lines, four LoRA experts, and fixed per-source exposure and per-expert budgets, a grouping derived purely from initial-state single-source probes and an uninformed uniform default were each trained once for one interval from a shared warm-up state and served by routing computed from their block allocations. Both scored below the label reference ($-2.43$ and $-2.54$ pp), neither showed the prespecified 2 pp gain, and in a post hoc comparison the probe-derived grouping was indistinguishable from the uniform default ($+0.11$ pp, $[-1.25,+1.54]$). The probes suggested structure that the tree does not encode, most clearly the upward effect of the other mathematics sources on Statistics (a descriptive pooled count), but a grouping built from them did not deliver the advantage predicted from the probes once carried to a later training state, a finite budget, and specified routing. These conclusions are conditional on one starting state, one interval, one routing rule, and one training run per candidate, and they use public data rather than business-line data. A further attempt to group by probes should, before training any candidate, measure the probes at the state where the decision is made, fix or record routing separately from allocation, and size the evaluation for sampled decoding; Section~\ref{sec:open} estimates what these and the other open measurements would cost.

\subsection*{AI use statement}
This work is AI-assisted; it is not autonomous AI research. The research question, the experimental decisions, and the interpretation of the results were proposed and adjudicated by the human authors. AI coding agents (Claude by Anthropic and Codex by OpenAI) were used extensively to write and test the code, run the analyses, assemble the evidence package, and draft and edit this report. The authors take responsibility for the full content of the report and the evidence package, including text, numbers, and code produced with AI assistance.

\appendix

\section{Probe statistics and multiplicity}\label{app:stats}

\paragraph{Probe design.}
A probe starts from $S_0$, the base model with a freshly initialized LoRA adapter ($B=0$), trains it on 16 blocks from one source, each with at least 32{,}768 completion tokens, and scores paired correctness on the audit split, 80 items for each of the seven sources (all from OpenCodeInstruct for the two code sources). Two training-stream shuffles were run. Shuffle s17 completed all seven training rows and s23 completed six; its generic-programming row is missing because the run was stopped before that row finished, and the six completed rows were recovered from the run log. The shuffles change only the order of training blocks and share the audit items and the initial predictions, so they are not independent item-level replications. $S_0$ accuracies are $0.3750$ (Integral), $0.4625$ (Differential), $0.3500$ (Abstract), $0.3875$ (Linear), $0.2000$ (Statistics), $0.3625$ (algorithmic), and $0.2000$ (generic). For a cell (training source $\to$ scored source), $+g/-l$ counts items that changed from incorrect to correct and from correct to incorrect, and $p$ is the two-sided exact McNemar test, a binomial test on the $g+l$ discordant items.

\begin{table}[t]
\centering\small
\caption{Probe cells with nominal $p<0.05$ in either shuffle, plus the boundary case Integral$\to$Abstract. Two-sided exact McNemar $p$; the Bonferroni-adjusted value is $p_{\rm Bonf}=\min(1,49p)$ over the 49-cell family; $^\dagger$ marks the only cell with $p_{\rm Bonf}<0.05$ ($p_{\rm Bonf}=0.00103$). The two shuffles share audit items and initial predictions.}
\label{tab:stats-cells}
\begin{tabular}{lrrrr}
\toprule
& \multicolumn{2}{c}{s17 (7 rows)} & \multicolumn{2}{c}{s23 (6 rows)}\\
\cmidrule(lr){2-3}\cmidrule(lr){4-5}
Cell & $+g/-l$ & $p$ & $+g/-l$ & $p$\\
\midrule
Linear $\to$ Statistics & $+16/-3$ & 0.004 & $+14/-4$ & 0.031\\
Statistics $\to$ Statistics & $+18/-4$ & 0.004 & $+13/-6$ & 0.167\\
Linear $\to$ Integral & $+21/-6$ & 0.006 & $+17/-10$ & 0.248\\
Statistics $\to$ Linear & $+22/-8$ & 0.016 & $+19/-8$ & 0.052\\
Abstract $\to$ Abstract & $+21/-9$ & 0.043 & $+20/-15$ & 0.500\\
Integral $\to$ Abstract & $+22/-10$ & 0.0501 & $+17/-11$ & 0.345\\
Differential $\to$ Statistics & $+12/-7$ & 0.359 & $+20/-1$ & $2.1\times10^{-5}$\,$^\dagger$\\
Linear $\to$ Linear & $+16/-10$ & 0.327 & $+24/-8$ & 0.007\\
Statistics $\to$ Abstract & $+20/-13$ & 0.296 & $+23/-10$ & 0.035\\
\bottomrule
\end{tabular}
\end{table}

\paragraph{Single cells and multiplicity.}
In s17, five of the 49 cells reach nominal $p<0.05$: Abstract$\to$Abstract, Linear$\to$Integral, Linear$\to$Statistics, Statistics$\to$Linear, and Statistics$\to$Statistics (Table~\ref{tab:stats-cells}), against at most 2.45 expected under a complete null ($49\times0.05$; about 1.35 for the exact test at the observed discordant counts, since the discrete test rejects with probability below 0.05). Integral$\to$Abstract ($p=0.0501$) is not among them. In s23, four of the 42 measured cells reach nominal $p<0.05$; Linear$\to$Statistics reproduces nominally ($+14/-4$, $p=0.031$), and Statistics$\to$Linear narrowly misses ($+19/-8$, $p=0.052$). Under a Bonferroni correction over the 49-cell family (conservative for the 42 cells of s23), no cell passes in s17; in s23, Differential$\to$Statistics passes ($p_{\rm Bonf}=0.00103$), and only that cell; no edge passes in both shuffles. Individual edges are therefore treated as hypotheses, not as confirmed effects.

\paragraph{Statistics column and the generic control.}
Aggregating by scored source rather than by cell, each of the four other mathematics sources produced more gains than losses on Statistics in both shuffles. Pooled over the four donors and the eight source--shuffle measurements, Statistics had $+111/-36$ discordant items (Integral $+22/-9$, Differential $+32/-8$, Abstract $+27/-12$, Linear $+30/-7$). Treating the eight measurements as exchangeable, which has not been verified, gives a nominal $p\approx4\times10^{-10}$; we report the aggregate as descriptive. The strongest donor differs between shuffles (Linear, $+16/-3$, in s17; Differential, $+20/-1$, in s23). Generic programming has the same $S_0$ accuracy as Statistics ($0.2000$), yet the same four donors moved it $+55/-60$ ($p\approx0.71$). The Statistics pattern is therefore not explained by baseline accuracy alone in this panel; this single comparison does not settle what does explain it.

\paragraph{The Linear--Statistics pair and selection bias.}
Pooled over both directions and both shuffles, Linear Algebra and Statistics have $+71/-23$ discordant items (75.5\% positive). The pair was singled out after inspecting the $7\times7$ tensor, so these counts carry selection bias and are descriptive. Compared with pairs that the label tree specifies, which were fixed before looking at the data, its positive fraction exceeds those of the two mathematics label pairs (Abstract--Linear $+66/-42$ and Integral--Differential $+60/-41$; two-sided Fisher exact $p=0.035$ and $0.022$) but not that of the code label pair (algorithmic--generic $+44/-22$, from both directions in s17 and one direction in s23; $p=0.284$). Because the pair was chosen after the fact, even the first two comparisons overstate the evidence. An independent confirmation of the Linear--Statistics coupling was planned but never run. C2's Linear Algebra plus Statistics expert came out of the transportation linear program over the whole s17 tensor, which contains these cells (Section~\ref{sec:probes}); the candidate in Appendix~\ref{app:c3} was built on the coupling directly.

\paragraph{Simultaneous intervals for per-line candidate differences.}
The candidate round yields 21 per-line differences $\Delta_\ell$ relative to C0: the three executed non-reference candidates, including the one in Appendix~\ref{app:c3}, times seven lines. Added after observing results, we bound them jointly with a max-$|t|$ bootstrap. In each of $B=2{,}000$ jointly paired resamples (items resampled within source, indices shared by all candidates) we record $t^*=\max_\ell|\Delta^*_\ell-\hat\Delta_\ell|/\widehat{\rm se}_\ell$, where $\widehat{\rm se}_\ell$ is the bootstrap standard deviation; the critical value $q$ is the $\lfloor0.95B\rfloor$-th order statistic of $t^*$ (zero-based), and the intervals are $\hat\Delta_\ell\pm q\,\widehat{\rm se}_\ell$. Including the appendix candidate makes the family larger, and the intervals for C1 and C2 wider, than a family of 14 would. With the post hoc bootstrap seed (20261009), $q=3.025$, against 1.96 for a single 95\% interval; over 22 bootstrap seeds, $q$ ranged from 2.965 to 3.094. The simultaneous interval for C1 on generic programming lay entirely below zero under every seed (upper limit $-0.82$ to $-0.29$ pp). For C2 on algorithmic programming the upper limit sat at zero: $-0.07$ pp under the post hoc seed, $+0.03$ pp when recomputed with the bootstrap seed of the primary analysis (20261007), and between $-0.07$ and $+0.23$ pp across the 22 seeds (below zero for 7 of them), so we treat that line as borderline. No simultaneous interval lies entirely above zero; the largest upper limits are $+6.62$ pp (C1, Abstract), $+6.45$ pp (C2, Statistics), and $+7.15$ pp (the appendix candidate, Abstract).

\section{Post hoc proxy diagnostic}\label{app:proxy}

This diagnostic was added after observing results. It asks how the additive model from which C2 was derived scores the executed candidates under several readings, and compares those scores with the measurements.

\paragraph{Model.}
Let $D_{k,j}$ be the $S_0$ probe change in accuracy on source $j$ after 16 blocks of source $k$, and $\beta_j$ the $S_0$ accuracy, both on the audit split. The model predicts the competence of expert $i$ after it receives $n_{ik}$ blocks of each source $k$, and the routed utility, as
\begin{equation}
\hat c_{i,j}=\operatorname{clip}_{[0,1]}\Bigl(\beta_j+\sum_kD_{k,j}\,\frac{\min(n_{ik},16)}{16}\Bigr),
\qquad \hat U=\frac17\sum_j\hat c_{h(j),j},
\label{eq:proxy-model}
\end{equation}
the formula used during development. Three readings differ in routing and transfer: \emph{block-matrix routing}, the executed rule $h=h_{\rm blk}$ (Eq.~\eqref{eq:blockrouting}); \emph{competence routing}, $h(j)=\arg\max_i\hat c_{i,j}$; and \emph{zero transfer}, which keeps only the diagonal terms $D_{k,k}$ under block-matrix routing. Tensor s17 is the complete first shuffle; s23* is the second shuffle with its missing generic row borrowed from s17. Three checks passed: the block-matrix routes computed by the diagnostic equal those used in evaluation for all four candidates; the competence-routed utility agrees with the development code to within $1.1\times10^{-16}$; and C2 equals 56 times the optimum of the transportation linear program for the s17 model under competence routing ($\hat U^\star=0.4426$).

\begin{table}[t]
\centering\small
\caption{Model-predicted and measured differences from C0 (pp); post hoc. The two rightmost columns concern the candidate outside the setting (Appendix~\ref{app:c3}). Model rows use the $S_0$ audit tensors; the measured row uses state W, 400 test items per source, and two samples per item.}
\label{tab:proxy-full}
\begin{tabular}{llrrrr}
\toprule
Reading & Tensor & C1$-$C0 & C2$-$C0 & C3$-$C0 & C3 $\Delta U_{\rm LS}$\\
\midrule
Block-matrix routing & s17 & $-2.23$ & $+2.01$ & $+1.22$ & $+4.57$\\
 & s23* & $-1.92$ & $+0.67$ & $+0.74$ & $+2.27$\\
Competence routing & s17 & $-3.90$ & $+0.90$ & $+0.58$ & $+0.94$\\
 & s23* & $-3.67$ & $+1.06$ & $+0.40$ & $+1.64$\\
Zero transfer & s17 & $-3.73$ & $+0.02$ & $-0.02$ & $-0.55$\\
 & s23* & $-2.98$ & $-0.10$ & $-0.02$ & $-0.27$\\
\midrule
Measured & & $-2.54$ & $-2.43$ & $-1.54$ & $-3.56$\\
\bottomrule
\end{tabular}
\end{table}

\begin{table}[t]
\centering\small
\caption{Model-predicted and measured differences from C0 on the Linear--Statistics lines ($\Delta U_{\rm LS}$) and the other five lines ($\Delta U_{\rm other5}$), pp; post hoc.}
\label{tab:proxy-split}
\begin{tabular}{llrrrr}
\toprule
& & \multicolumn{2}{c}{C1$-$C0} & \multicolumn{2}{c}{C2$-$C0}\\
\cmidrule(lr){3-4}\cmidrule(lr){5-6}
Reading & Tensor & $\Delta U_{\rm LS}$ & $\Delta U_{\rm other5}$ & $\Delta U_{\rm LS}$ & $\Delta U_{\rm other5}$\\
\midrule
Block-matrix routing & s17 & $+0.23$ & $-3.22$ & $+4.92$ & $+0.84$\\
 & s23* & $-1.48$ & $-2.09$ & $+1.95$ & $+0.16$\\
Competence routing & s17 & $-3.40$ & $-4.09$ & $+1.29$ & $+0.75$\\
 & s23* & $-3.75$ & $-3.64$ & $+3.12$ & $+0.23$\\
Zero transfer & s17 & $-4.45$ & $-3.44$ & $-0.23$ & $+0.13$\\
 & s23* & $-4.77$ & $-2.27$ & $-0.62$ & $+0.11$\\
\midrule
Measured & & $-1.31$ & $-3.02$ & $-1.13$ & $-2.95$\\
\bottomrule
\end{tabular}
\end{table}

\paragraph{Readings.}
Under block-matrix routing the model placed C2 above C0 with both tensors ($+2.01$ and $+0.67$ pp) and C1 below it ($-2.23$ and $-1.92$ pp); measured, both were below C0 (Table~\ref{tab:proxy-full}). The prediction for C1 has the measured sign, and the prediction for C2 does not. The largest predicted C2 advantage is on the Linear--Statistics lines ($+4.92$ and $+1.95$ pp), where the measurement was $-1.13$ pp (Table~\ref{tab:proxy-split}). With every transfer term removed, the model puts C2 level with C0 ($+0.02$ and $-0.10$ pp). Within this model, therefore, the explanation that probed transfer vanished by state W is insufficient to account for C2 scoring below C0. Switching C0 alone from block-matrix to competence routing raises its model utility by $1.66$ pp (s17) and $1.75$ pp (s23*), so in the model the routing rule matters on the same scale as the differences being explained.

\paragraph{Limits.}
The diagnostic cannot show that the measured gap comes from joint-training interference or from any other single mechanism. Besides the routing, which the readings vary, and the training mix, for which the model has no term, the model and the measurement differ in state ($S_0$ versus W), item bank (80 audit items versus 400 test items per source; for algorithmic programming, OpenCodeInstruct audit items versus test items that are mostly from MBPP+), and dose (16 blocks of one source versus 0--8 blocks of each source within a mixed interval). C2 is the optimum of the same s17 model, so its predicted advantage is inflated by selection (the optimizer's curse), and s23* shares items, initial predictions, and one row with s17, so it is not an independent check. Only three candidates, two of them inside the setting, are compared with C0; three points cannot establish a general relation between predicted and measured differences.

\section{Evaluation protocol and decoding variance}\label{app:protocol}

\paragraph{Protocol.}
The protocol was frozen before any candidate was trained. It fixed the candidates, the starting state, the routing rule, the evaluation items and seeds, the primary metric, $\delta_{\min}$, and the reading rule, and it excluded adding candidates or items after results were seen. Each candidate was trained from W for one interval of 56 blocks with fixed allocation (about 4{,}570~s each). Routing followed the block-matrix rule, and no route calibration was run. Evaluation used the first 400 items of each source's test split in a fixed sorted order (the smallest test bank has 421 items), the same items for every candidate. The algorithmic test split holds the 376 MBPP+ items \citep{liu2023evalplus} and 768 OpenCodeInstruct held-out items, and its first 400 in sorted order are the 376 MBPP+ items and 24 OpenCodeInstruct items; for this source the move from the audit split, on which it was probed, to the test split is therefore also a change of benchmark. Each item was answered twice at temperature $T=0.8$, with CUDA seeds 2001 and 2002 for every candidate, with at most 8{,}192 new tokens for mathematics and 1{,}024 for code. Each candidate--seed evaluation (2{,}800 generations) took about 32{,}300~s. An item's score is the mean over its samples, and per-source accuracy is the mean over items; $U_{\rm all}$ is the uniform average over the seven sources, $U_{\rm LS}$ the average over Linear Algebra and Statistics, and $U_{\rm other5}$ the average over the other five. The primary metric is $\Delta U_{\rm all}$ relative to C0; $\Delta U_{\rm LS}$, $\Delta U_{\rm other5}$, and per-source differences were also to be reported.

\paragraph{Intervals.}
All intervals are jointly paired percentile bootstrap intervals with $B=2{,}000$ resamples: within each source, items are resampled with replacement, and the same indices are used for every candidate. The endpoints are the order statistics $v_{(\lfloor0.025B\rfloor)}$ and $v_{(\lfloor0.975B\rfloor)}$ (zero-based) of the resampled values; the primary analysis uses seed 20261007 and the post hoc analyses seed 20261009. Intervals cover test-item and decoding sampling only, conditional on one training run per candidate. Per-source differences are exact multiples of $0.125$ pp ($1/800$), so we report per-source values to three decimals.

\paragraph{Minimum useful gain and reading rule.}
The minimum useful gain $\delta_{\min}=2$ pp on $U_{\rm all}$ was set before any candidate result, from measurement capability rather than from an expected effect. With 400 items per source and one sample per item, the paired standard error of $\Delta U_{\rm all}$ between the two four-expert portfolios of the earlier comparison described under the choice of $k$ below, both trained from W, was 0.98 pp (only the variance of their per-item difference was used, not the difference itself), so the minimum detectable effect at 80\% power was about 2.7 pp. True gains below 2 pp are thus outside what this design can find. The frozen reading rule maps the 95\% interval of $\Delta U_{\rm all}$ for each candidate to one of three readings: lower limit above $+\delta_{\min}$, an opportunity worth pursuing; interval inside $(-\delta_{\min},+\delta_{\min})$, the candidate is unlikely to provide the set gain at this state; interval crossing $\pm\delta_{\min}$, insufficient precision. The script \path{analysis/analyze_candidates.py} also has a fourth reading, interval entirely below $-\delta_{\min}$, which was not part of the frozen rule; no candidate interval falls in it. Every candidate interval crosses $-\delta_{\min}$, so the literal reading is insufficient precision. Because the rule was written for gains, the main text answers three questions separately: a $+2$ pp gain is not supported, since every upper limit is below zero; a shortfall relative to C0 is supported; whether the shortfall reaches 2 pp is not resolved. Neither the rule nor the statistics were changed.

\paragraph{Choice of $k$.}
The rule for the number of samples per item was written before the variance decomposition: choose the smallest $k\in\{1,2,4\}$ whose minimum detectable effect for $\Delta U_{\rm all}$ (80\% power, two-sided $\alpha=0.05$, test sampling only) is at most $\delta_{\min}$, or $k=4$ with the shortfall recorded. The decomposition used two independent samples (the default CUDA generator state and seed 12345) of one checkpoint on the same 80 test items per source. That checkpoint was a four-expert portfolio trained from W for one interval in an earlier comparison that is not part of this report; the evidence package contains only the summary numbers (\path{results/protocol/decoding_variance_summary.json}), not the per-item scores, the script that computed the summary, or the training code of the portfolios involved. For algorithmic programming all 80 items were MBPP+ items, so that row of Table~\ref{tab:protocol-decoding} describes MBPP+. Per source, the decoding variance is $\sigma_d^2=\mathbb E[(y_1-y_2)^2]/2$, the single-sample variance $\sigma_t^2$ is the mean over the two samples, and the between-item variance is $\sigma_b^2=\sigma_t^2-\sigma_d^2$ (Table~\ref{tab:protocol-decoding}). Two samples disagreed on 21--34\% of items, and decoding accounted for 42--67\% of single-sample variance. The per-item variance of a paired difference with $k$ samples was extrapolated as $V(k)=V_{\rm obs}-2\sigma_d^2(1-1/k)$ from the observed per-item variance $V_{\rm obs}$ of a 400-item paired difference at $k=1$ between this portfolio and a second four-expert portfolio of the same earlier comparison, also trained from W for one interval, giving standard errors of 0.98, 0.69, and 0.48 pp and minimum detectable effects of 2.74, 1.93, and 1.36 pp for $k=1,2,4$. The rule selected $k=2$, and seeds 2001 and 2002 were then fixed. The extrapolation assumes that decoding noise is independent between arms. In the candidate round, the two samples of a candidate disagreed on 20.5--34.0\% of items per source for C0--C2 and on 19.5--33.0\% for the candidate of Appendix~\ref{app:c3}.

\begin{table}[t]
\centering\small
\caption{Decoding-variance decomposition used to choose $k$: two independent samples of one model on 80 items per source. The 80 algorithmic items are MBPP+ problems.}
\label{tab:protocol-decoding}
\begin{tabular}{lrrrrr}
\toprule
Source & Disagreeing items & $\sigma_d^2$ & $\sigma_t^2$ & $\sigma_b^2$ & Decoding share\\
\midrule
Integral & 22/80 & 0.138 & 0.252 & 0.115 & 54\%\\
Differential & 27/80 & 0.169 & 0.253 & 0.084 & 67\%\\
Abstract & 25/80 & 0.156 & 0.248 & 0.092 & 63\%\\
Linear & 23/80 & 0.144 & 0.252 & 0.108 & 57\%\\
Statistics & 20/80 & 0.125 & 0.237 & 0.112 & 53\%\\
algorithmic & 17/80 & 0.106 & 0.250 & 0.144 & 42\%\\
generic & 17/80 & 0.106 & 0.248 & 0.142 & 43\%\\
\bottomrule
\end{tabular}
\end{table}

\paragraph{Interval-control diagnostic.}
This diagnostic was added after observing results and is used only as a diagnostic. In C0 and in the candidate of Appendix~\ref{app:c3}, experts e1 and e2 received identical blocks and served the same four sources (Integral, Differential, algorithmic, generic). We call these training-allocation- and routing-matched control sources. Their final parameters are not strictly identical: for e1 and e2 the parameter distance between the two runs was about 0.027, against a training displacement from W of about 1.4--1.6 (about 2\%, from GPU nondeterminism), whereas e0 and e3 differed by 0.88--0.94. Their true expected performance therefore need not be equal. The differences on these sources were $-3.625$ $[-7.500,+0.375]$, $-0.875$ $[-4.625,+2.875]$, $+0.875$ $[-2.125,+3.875]$, and $-1.500$ $[-4.500,+1.500]$ pp; every interval contains zero, and these controls showed no evident anomaly. Four intervals that contain zero cannot verify 95\% coverage, and the controls are not replicates of a known null, so this diagnostic does not establish the coverage of the per-source intervals. On these sources, per-item agreement between C0 and the appendix candidate was 69--88\% when both used the same sampling seed, close to the 66--82\% when they used different seeds; no evident per-item agreement advantage from common random numbers was observed. The variance benefit of common random numbers depends on the covariance between the two arms' outputs, not on token-level identity, so this does not show that decoding noise is independent between arms, and the independence assumed in choosing $k$ remains an assumption. On sources whose training and routing were matched, single-source point differences reached about 3.6 pp under this evaluation. We do not attribute that difference to evaluation noise alone, but it calls for caution in reading single-source differences.

\section{An additional executed candidate outside the setting}\label{app:c3}

\paragraph{Construction and why it is outside the setting.}
The frozen protocol included a fourth candidate, C3, which edits the label reference. Starting from C0, it swaps Abstract Algebra and Statistics between experts e0 and e3, forming a Linear Algebra plus Statistics expert. Its blocks, with rows for experts and columns for Integral, Differential, Abstract, Linear, Statistics, algorithmic, and generic, are e0 $(0,0,0,7,7,0,0)$, e1 $(7,7,0,0,0,0,0)$, e2 $(0,0,0,0,0,7,7)$, and e3 $(1,1,8,1,1,1,1)$, and its block-matrix routing is $(1,1,3,0,0,2,2)$. The edit followed the development hypothesis of a Linear--Statistics coupling (Appendix~\ref{app:stats}); it is not a solution of the transportation linear program. Of its other groups, the calculus expert and the code expert are copied verbatim from C0, and the Abstract-dominated expert is C0's Statistics expert, one-block filler included, with Abstract Algebra in place of Statistics; all three therefore require the label tree. In the setting of this report, where the standard grouping is unknown, C3 cannot be constructed, so it is not a candidate for the report's question and is excluded from the main comparison. Classifying it as outside the label-unknown setting was decided after observing results, when the report was reframed around that setting. C3 was specified, trained, and scored under the same frozen protocol as the other candidates, and we report it in full so that no executed arm is omitted. Inside the setting, the Linear--Statistics pairing is tested by C2, whose expert e0 trains on 6 Linear Algebra and 8 Statistics blocks and serves both lines.

\paragraph{Results.}
C3 had $U_{\rm all}=0.4882$, $U_{\rm LS}=0.4662$, and $U_{\rm other5}=0.4970$; its two samples disagreed on 19.5--33.0\% of items per source. Against C0, a comparison specified before the run, its overall utility was lower by $1.54$ pp $[-2.89,-0.20]$ (Table~\ref{tab:c3-main}). The difference was largest on the two lines the edit regrouped, $\Delta U_{\rm LS}=-3.56$ pp, with the other five lines at $-0.72$ pp. Under the frozen reading rule its interval crosses $-\delta_{\min}$ (insufficient precision); its upper limit is below zero, so a $+2$ pp gain is not supported. Against C1, a comparison added after observing results, C3's overall difference was $+1.00$ pp $[-0.37,+2.45]$; the upper limit exceeded $+2$ pp under each of 22 bootstrap seeds (from $+2.30$ to $+2.50$), so an advantage of the prespecified size over C1 is not excluded. On the five lines other than Linear Algebra and Statistics, four of which are served by experts trained on the same blocks as C0's, C3 exceeded C1 by $+2.30$ pp $[+0.67,+4.02]$, and on Linear Algebra and Statistics it trailed C1 by $2.25$ pp $[-5.00,+0.50]$; these post hoc intervals are not adjusted for multiplicity.

\begin{table}[t]
\centering\small
\caption{The candidate outside the setting, C3: differences in pp with 95\% jointly paired bootstrap intervals. The comparison with C0 was specified before the run; the comparison with C1 was added after observing results and is not adjusted for multiplicity.}
\label{tab:c3-main}
\begin{tabular}{lrrr}
\toprule
& $\Delta U_{\rm all}$ & $\Delta U_{\rm LS}$ & $\Delta U_{\rm other5}$\\
\midrule
C3 $-$ C0 & $-1.54$ $[-2.89,-0.20]$ & $-3.56$ $[-6.25,-0.88]$ & $-0.72$ $[-2.30,+0.83]$\\
C3 $-$ C1 & $+1.00$ $[-0.37,+2.45]$ & $-2.25$ $[-5.00,+0.50]$ & $+2.30$ $[+0.67,+4.02]$\\
\bottomrule
\end{tabular}
\end{table}

\begin{table}[t]
\centering\small
\caption{Per-line differences C3 $-$ C0 (pp). Point estimates are exact multiples of 0.125 pp. Marginal 95\% intervals come from the analysis specified before the run; simultaneous 95\% intervals (max-$|t|$ bootstrap over all 21 executed per-line comparisons, $q=3.025$; Appendix~\ref{app:stats}) were added after observing results.}
\label{tab:c3-lines}
\begin{tabular}{lrrr}
\toprule
Line & Difference & Marginal 95\% & Simultaneous 95\%\\
\midrule
Integral & $-3.625$ & $[-7.500,+0.375]$ & $[-9.43,+2.18]$\\
Differential & $-0.875$ & $[-4.625,+2.875]$ & $[-6.78,+5.03]$\\
Abstract & $+1.500$ & $[-2.125,+5.000]$ & $[-4.15,+7.15]$\\
Linear & $-4.750$ & $[-8.625,-0.875]$ & $[-10.64,+1.14]$\\
Statistics & $-2.375$ & $[-6.125,+1.125]$ & $[-7.80,+3.05]$\\
algorithmic & $+0.875$ & $[-2.125,+3.875]$ & $[-3.75,+5.50]$\\
generic & $-1.500$ & $[-4.500,+1.500]$ & $[-5.96,+2.96]$\\
\bottomrule
\end{tabular}
\end{table}

\paragraph{Traffic sensitivity and per-line resolution.}
Because utility is linear in traffic weights, any reweighting places C3 $-$ C0 in $[-4.750,+1.500]$ pp, from all traffic on Linear Algebra to all traffic on Abstract Algebra, so no reweighting gives a point-estimate advantage of $\delta_{\min}$. No simultaneous interval lies entirely above or below zero (Table~\ref{tab:c3-lines}). The largest simultaneous upper limit is $+7.15$ pp (Abstract), so a large gain for traffic concentrated on Abstract Algebra is not excluded. The marginal interval for Linear Algebra lies below zero, but it is one of 21 per-line comparisons, and its simultaneous interval contains zero.

\paragraph{Proxy predictions.}
The additive model of Appendix~\ref{app:proxy} predicted C3 $-$ C0 of $+1.22$ pp (s17) and $+0.74$ pp (s23*) under block-matrix routing, $+0.58$ and $+0.40$ pp under competence routing, and $-0.02$ pp for both tensors with transfer removed, against the measured $-1.54$ pp. For the Linear and Statistics lines it predicted $\Delta U_{\rm LS}$ of $+4.57$ and $+2.27$ pp under block-matrix routing ($+0.94$ and $+1.64$ with competence routing; $-0.55$ and $-0.27$ without transfer), against the measured $-3.56$ pp. Unlike C2, C3 was not selected by maximizing the model, but the pair it builds on was selected after inspecting the tensor. The limits stated in Appendix~\ref{app:proxy} apply. A retraining of C0 and C3 has not been run; at the unit costs of Section~\ref{sec:open} it would take about $2\times19$ GPU-hours (per candidate, one training interval of about 4{,}570\,s plus two evaluations of about 32{,}300\,s each), and if it is run, it should be reported as a new comparison, not pooled with this round.

\section{Evidence provenance}\label{app:provenance}

% Project repository supplied by the authors; URL inserted without verification of its contents.
\paragraph{Evidence package.}
Every experimental number in this report is read from a file of the evidence package \path{transfer-probes-evidence/} (\url{https://github.com/CodePothunter/Reproduction-for-From-Transfer-Probes-to-Expert-Grouping}), set in its code and launch scripts (base model, LoRA rank, sampling temperature, block size, generation limits, and seeds), or computed from such numbers by a formula stated in the text (Section~\ref{sec:open}). Paths below are relative to the package root.
\begin{itemize}
\item Data manifest: built by \path{pipeline/build_manifest_7src.py}; split counts, the composition of the algorithmic test split (376 MBPP+ and 768 OpenCodeInstruct items), the first 400 test items per source, and the manifest's SHA-256 are in \path{results/protocol/manifest_summary.json}. The manifest itself, which contains the problem texts, is not included.
\item Probes: shuffle s17, \path{results/probes/s17_transfer.json} (seven rows); shuffle s23, \path{results/probes/s23_transfer_partial.json} (six rows, recovered from the run log by \path{pipeline/salvage_partial_tensor.py}). Probe code: \path{pipeline/probe_transfer.py}; launch script \path{pipeline/run_probes.sh}, whose arguments are inferred from the run logs because the command lines of the two runs were not recorded.
\item Probe statistics (cell tests, Bonferroni correction, expected counts under the null, column and pair counts, Fisher exact tests in exact integer arithmetic): \path{analysis/probe_stats.py}, output \path{results/analysis/PROBE-stats.json}.
\item Label-tree cuts, the construction of C0, and its tie in the tree-distance transportation program: \path{analysis/c0_construction.py}, output \path{results/analysis/C0-construction.json}. Development-time proxy sweep: \path{analysis/allocation_sweep.py}, output \path{results/analysis/ALLOC-sweep.txt}.
\item Warm-up state W and candidate training: \path{pipeline/run_pilot.py} with the launch scripts \path{pipeline/run_warmup.sh} and \path{pipeline/run_candidates.sh}; the block matrix and token counts of W, and the block matrices, token counts, routes, and wall-clock times of C0--C3, are in \path{results/candidates/training_summary.json}. The run that produced W also executed measurement-only routines that are not part of this report; that run asserted each time that they left the training state unchanged (its log is not included), and \path{pipeline/run_warmup.sh} omits them. On small CPU scenarios, \path{tests/test_strip_equivalence.py} checks that the package code reproduces the warm-up, candidate-training, probe, and evaluation trajectories of the unreduced code that trained and evaluated the candidates. The probes and W were run with earlier revisions of that code; their differences from it lie outside the probe path and do not change the blocks or the training of the uniform warm-up interval.
\item Test evaluation: \path{pipeline/score_test_extension.py} and \path{pipeline/run_eval.sh}; per-item scores of every candidate and seed in \path{results/candidates/C*-eval-s2001.json} and \path{results/candidates/C*-eval-s2002.json}, with item identifiers replaced by truncated SHA-256 hashes of the manifest identifiers.
\item Primary analysis: \path{analysis/analyze_candidates.py}, output \path{results/analysis/CANDIDATES-analysis.json}.
\item Post hoc analyses: \path{analysis/analyze_label_reference.py}, output \path{results/analysis/LABELREF-analysis.json}; \path{analysis/proxy_posthoc.py}, output \path{results/analysis/PROXY-posthoc.json}; per-item agreement on the control sources, \path{analysis/control_agreement.py}, output \path{results/analysis/CONTROL-agreement.json}; parameter distances, \path{analysis/param_distance.py}, output \path{results/analysis/PARAM-distance.json}.
\item Decoding variance and the choice of $k$: \path{results/protocol/decoding_variance_summary.json} (summary only; see Appendix~\ref{app:protocol}).
\end{itemize}
The script \path{analysis/run_all.sh} recomputes every output under \path{results/analysis/} on CPU from the files under \path{results/}, except the parameter distances.

\paragraph{What the package cannot reproduce.}
The model checkpoints of W and of the candidates, and the data manifest, are too large to include; regenerating them requires the GPU pipeline above. GPU kernels are not deterministic, so a rerun reproduces the data order, the block allocations, and the routes, but not bit-identical parameters or scores (Appendix~\ref{app:protocol}). The inputs of the parameter-distance analysis (three checkpoints, identified by SHA-256 in its output), and the inputs of the decoding-variance summary and the script that computed it, are not included. The frozen protocol document is not included either. Its machine-checkable parts are in the package: $\delta_{\min}$, the reading rule (with the additional fourth reading noted in Appendix~\ref{app:protocol}), the contrasts against C0, and the bootstrap scheme in \path{analysis/analyze_candidates.py}; the candidate block matrices, the routing rule, the evaluation items, $k$, and the evaluation seeds in \path{pipeline/run_candidates.sh} and \path{pipeline/run_eval.sh}.

\end{document}